\documentclass[lettersize,journal]{IEEEtran}
\usepackage{amsmath,amsfonts}
\usepackage{algorithmic}
\usepackage{algorithm}
\usepackage{array}
\usepackage[caption=false,font=normalsize,labelfont=sf,textfont=sf]{subfig}
\usepackage{textcomp}
\usepackage{multirow}
\usepackage{stfloats}
\usepackage{url}
\usepackage{xcolor}
\usepackage{verbatim}
\usepackage{graphicx}
\usepackage{cite}
\usepackage{epstopdf}
\usepackage{booktabs}
\begin{document}

\title{Reconstructing Deep Neural Networks: Unleashing the Optimization Potential of Natural Gradient Descent}

\author{Weihua Liu$\dagger$, Said Boumaraf$\dagger$, Jianwu Li, Chaochao Lin, Xiabi Liu\textsuperscript{*}, Lijuan Niu\textsuperscript{*}, Naoufel Werghi
\thanks{Weihua Liu is with the School of Medical Technology, Beijing Institute of Technology, Beijing 100081, China (e-mail: liuweihua@bit.edu.cn) and also serves as a researcher at AthenaEyesCO., LTD., Changsha, Hunan 410205, China.}
\thanks{Said Boumaraf and Naoufel Werghi are with the Department Computer Science, Khalifa University of Science and Technology, Abu Dhabi, United Arab Emirates (email: said.boumaraf@ku.ac.ae, naoufel.werghi@ku.ac.ae).}
\thanks{Jianwu Li, Chaochao Lin, and Xiabi Liu are with the School of Computer Science
\& Technology, Beijing Institute of Technology, Beijing 100081, China (email: ljw@bit.edu.cn, linchaochao@bit.edu.cn, liuxiabi@bit.edu.cn).}
\thanks{Lijuan Niu is with the department of Ultrasound National Cancer Center/National Clinical Research Center for Cancer/Cancer Hospital, Chinese Academy of Medical Sciences and Pekin Union Medical College, No. 17 Nanli Panjiayuan, Chaoyang 100021, Beijing China (email: niulijuan8197@126.com).}
\thanks{\textsuperscript{*} Corresponding authors: Xiabi Liu; Lijuan Niu}
\thanks{$\dagger$ Weihua Liu and Said Boumaraf contributed equally to this work.}
}



\maketitle

\begin{abstract}
Natural gradient descent (NGD) is a powerful optimization technique for machine learning, but the computational complexity of the inverse Fisher information matrix limits its application in training deep neural networks. To overcome this challenge, we propose a novel optimization method for training deep neural networks called structured natural gradient descent (SNGD). Theoretically, we demonstrate that optimizing the original network using NGD is equivalent to using fast gradient descent (GD) to optimize the reconstructed network with a structural transformation of the parameter matrix. Thereby, we decompose the calculation of the global Fisher information matrix into the efficient computation of local Fisher matrices via constructing local Fisher layers in the reconstructed network to speed up the training. Experimental results on various deep networks and datasets demonstrate that SNGD achieves faster convergence speed than NGD while retaining comparable solutions. Furthermore, our method outperforms traditional GDs in terms of efficiency and effectiveness. Thus, our proposed method has the potential to significantly improve the scalability and efficiency of NGD in deep learning applications. Our source code is available at https://github.com/Chaochao-Lin/SNGD.
\end{abstract}


\section{Introduction}

Deep neural networks have achieved unprecedented success in a myriad of complex tasks, establishing themselves as a cornerstone in the advancement of machine learning and artificial intelligence. Despite their success, the optimization of these networks remains a formidable challenge, critically influencing their performance and applicability. Traditional Gradient Descent (GD) methods, while being straightforward and computationally efficient, primarily leverage first-order information\cite{huang2017centered}, often resulting in slow convergence rates and sub-optimal performance in the presence of highly non-linear and non-convex objective functions\cite{amid2022locoprop}. In contrast, Natural Gradient Descent (NGD)\cite{amari1998natural} emerges as a robust second-order optimization technique, offering faster and more effective convergence by considering the underlying geometry of the parameter space through the Fisher information matrix. However, the computational intensity required to calculate and invert this matrix poses significant challenges, rendering NGD impractical for training deep neural networks at scale.

To surmount these obstacles, we propose Structured Natural Gradient Descent (SNGD), a novel optimization framework that preserves the advantages of NGD while significantly reducing its computational burden. The cornerstone of SNGD is its ability to deconstruct the global Fisher information matrix into manageable, local Fisher matrices, thereby simplifying the overall computation and making it more amenable to deep network training. This methodological innovation paves the way for exploiting the optimization potential of NGD across a wider array of deep learning applications without being hindered by prohibitive computational costs. The theoretical innovation underpinning SNGD is articulated through a comprehensive paradigm that establishes a novel equivalence between the optimization of the original network via NGD and the application of a more computationally efficient Gradient Descent (GD) on a strategically transformed network. This transformation is achieved by reconstructing the deep network in such a way that aligns the traditional gradient descent's update direction with that of the natural gradient, albeit with significantly reduced computational complexity and storage demands. Central to this transformative process is the introduction of the local Fisher layer- a novel architectural element ingeniously designed to encapsulate the local curvature information of the loss function space. By integrating these local Fisher layers into the network, SNGD imposes a structured constraint on the transformation of model parameters, thereby ensuring that each gradient update is informed by a precise understanding of the local geometric properties of the loss landscape. This results in stable and fast gradient updates, leading to faster parameter convergence speed and better performance compared to conventional first-order optimizers. We evaluate the performance of SNGD across a spectrum of deep neural network architectures, including multi-layer Perceptron (MLP), convolutional neural networks (CNN), long short-term memory networks (LSTM), and residual networks (ResNet), leveraging diverse datasets like MNIST, CIFAR-10, ImageNet, and Penn Treebank. Experimental results demonstrate that SNGD significantly outperforms traditional first-order optimizers, achieving faster convergence and better performance in various training tasks.

In summary, our contributions are threefold: (1) We delineate a comprehensive theoretical framework that elucidates the equivalence between GD and NGD through a transformative reconstruction of deep networks, offering a new perspective on optimizing deep learning models with reduced computational overhead. (2) We introduce the concept of the Local Fisher Layer, a novel mechanism designed to capture essential curvature information, thereby facilitating significant improvements in the speed and stability of parameter convergence. (3) Through rigorous experimental validation, we demonstrate the practical efficacy and adaptability of SNGD, underscoring its superiority in enhancing the training efficiency of deep neural networks across various architectures and datasets.

\section{Related Works}
In this section, we focus on NGD and address numerical challenges in matrix root/inverse computations by drawing inspiration from the normalization method. Our work sheds new light on NGD and regularization methods. 
\subsection{Natural gradient descent}
NGD was first proposed by Amari et al. \cite{amari1998natural} in 1998. The natural gradient is the direction of the fastest decrease of the error defined in the parameter space in Riemannian space. Amari's experiments show that NGD has a faster convergence rate than stochastic gradient descent (SGD). However, implementing NGD in deep neural networks is usually challenging due to the large corresponding Fisher matrix. 

To simplify the calculation of natural gradients, Bastian et al. \cite{bastian2011simplified} considered every two layers of the neural network as a model, and the resulting information matrix is called the block information matrix.  Martens et al. \cite{martens2015optimizing} proposed an effective method to implement the natural gradient algorithm called the approximate curvature of Kronecker coefficients (K-FAC), an approximate natural gradient algorithm for optimizing the cascading structure of neural networks. Zhang et al. \cite{zhang2019fast} theoretically analyzed the convergence rate of NGD on nonlinear neural networks based on square error loss, determined two prerequisites for effective convergence after random initialization, and proved that although K-FAC can converge to global optimization as long as it satisfies the prerequisites. George et al. \cite{george2018fast} proposed a novel approximate natural gradient method that is less computationally expensive than K-FAC. Their method uses diagonal variance on the basis of Kronecker eigenvectors. The characteristic of George and other methods is that diagonal variance is based on Kronecker eigenvectors.
Bernacchia et al. \cite{bernacchia2018exact} derived the exact expression of the natural gradient in deep linear networks, which exhibits morbid curvature similar to that in the nonlinear case. Given the complexity of deep neural networks, Sun et al. \cite{sun2017relative}  decomposed a neural network model into a series of local subsystems, based on which they defined the relative Fisher information metric, reducing the complexity of the optimization calculation of the whole network model.  Lin et al. \cite{lin2021tractable} further addressed the computational difficulty of the Fisher matrix by using local parameter coordinates based on covariances of Gaussian and Wishart-based distributions. Yang et al. \cite{yang2022sketch} designed randomized techniques for networks and low-rank approximations to reduce the computational cost and memory requirement of computing the natural gradient.

On a broader perspective, the computational challenge of Natural Gradient Descent (NGD) arises from its reliance on second-order statistics/derivation,  a fundamental issue extending across various deep learning models. For instance, in Graph Neural Networks, Wang et al. \cite{wangpami23} introduced bilinear mapping and attentional second-order pooling to address this problem. The work of \cite{tyagi20}  proposed an adaptive second-order tanning by adapting the multiple optimal learning factor MOLF \cite{ malalur15} to develop efficient automated training for standalone MLP classifiers. In his substantial analysis of convex optimization, Lee et al. \cite{lee15} tackled the bottleneck associated with the inversion matrix concerning second-order derivatives. They employed weighted analytic center frameworks and incorporated regularization terms into the objective function to constrain the amplitude of the Hessian.

\subsection{Normalization method}
Normalization techniques have been widely used in deep neural networks to improve training efficiency and generalization performance. In this subsection, we discuss the normalization method used in our work, which builds upon the idea that the effect of the natural gradient in the parameter space is similar to the whitening of the signal \cite{sohl2012natural}. Our approach is to add additional layers to re-express the parameter space of each layer so that the optimization can perform gradient descent in Riemann space. We also provide an overview of other popular normalization methods as follows.

Local response normalization is proposed in AlexNet \cite{krizhevsky2017imagenet}, while batch normalization (BN) \cite{ioffe2015batch} has become a popular choice for normalizing data along the batch dimension. However, BN has no significant effect when the batch dimension is small, and to avoid using it, new normalization methods have been proposed. layer Normalization (LN) \cite{ba2016layer} normalizes along the channel dimension, while instance normalization (IN) \cite{ulyanov2016instance} performs a BN-like operation, but only for each sample. Weight normalization (WN) \cite{salimans2016weight} recommends standardizing the parameter weights of each layer rather than the features.  Group normalization (GN) \cite{wu2018group} divides the channels into groups, not affected by the batch size, and spectrum normalization \cite{miyato2018spectral} is a new weight standardization technique used to stably train generative adversarial networks. Spatial adaptive normalization \cite{park2019semantic} is another approach that has advantages in terms of visual fidelity and alignment with the input layout when the semantic input layout is given.  To improve the effectiveness of normalization, Singh et al. \cite{singh2022feature} proposed a novel Feature Wise Normalization approach that improves the generalization and superiority of the normalization method. Similarly, Fan et al. \cite{fan2021adversarially} proposed an adaptive standardization and rescaling normalization to learn standardization and rescaling statistics from data of different domains, which improves the model's generalization performance.

\section{Structured natural gradient descent (SNGD)}
\begin{figure*}
  \centering
  \includegraphics[width=\textwidth]{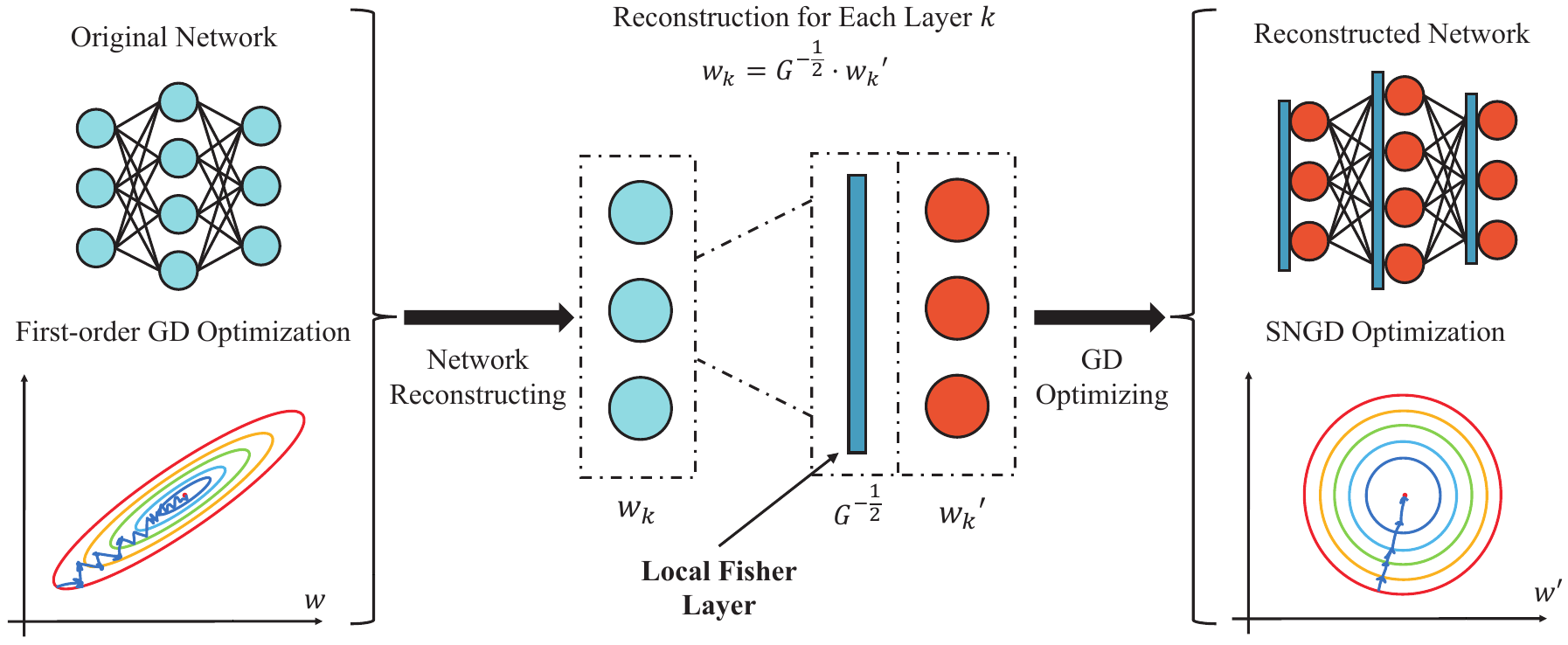}
  \caption{Overview of structured natural gradient descent (SNGD). The proposed method first reconstructs the original network with weights $w$ by the local Fisher layers, which correspond to Fisher information matrices. After that, GD is used to optimize the reconstructed network on new weights $w'$, leading to an approximation to the effect of NGD.}\label{fig1}
\end{figure*}

The structured natural gradient method (SNGD) is a technique that approaches the power of natural gradient descent (NGD) by reconstructing the original deep network and using traditional gradient descent (GD) optimization, which can make gradient updates more stable and faster compared to traditional first order GD. As shown in Fig. \ref{fig1}, our method utilizes decomposition to divide deep networks into subsystems hierarchically containing local Fisher layers and then use traditional GD to update the parameters of the reconstructed network. In this section, we first describe the relationship between NGD and GD, followed by the proposal of our SNGD and a description of how it can be applied to deep networks.

\subsection{Background}
NGD \cite{amari1998natural} is a second-order optimization method for training statistical models. The update rules for the model parameter vector $w$ are as follows:
\begin{equation}\label{eq1}
w^{(k+1)}\leftarrow w^{(k)}- \alpha \cdot G^{-1} \cdot \frac{\partial l}{\partial w^{(k)} }
\end{equation}
where $l$ is the loss function, $\frac{\partial l}{\partial w^{(k)} }$ is the gradient of the loss function calculated in $k$-th iteration, $\alpha$ is the learning rate, $G$ is the Fisher matrix, which can be regarded as the Riemann metric on the statistical manifold, and $G^{-1}$ is the inverse of the Fisher matrix. 

Assuming that the probability distribution of the model is $p(x)$ of the input $x$, the Fisher matrix $G$ is given by:
\begin{equation}\label{eq2}
G \leftarrow {\rm E}_{p(x|w)}\{\nabla_w \log{p(x|w)}{[\nabla_w \log{p(x|w)}]}^T \}
\end{equation}
where ${\rm E}_{p(x|w)} [\cdot]$ denotes the expectation with respect to probability distribution $p(x|w)$, $\nabla_w \log{p(x|w)}$ is the conventional gradient of the log-likelihood, and $[\nabla_w p(x|w)]^T$ is the transpose of the gradient matrix.

\subsection{Relationship between the NGD and GD}
We first discuss the relationship between NGD and GD and give a proposition.

\paragraph{Proposition 1} Let $w'=G^{\frac{1}{2}} \cdot w$. Then optimizing parameters $w'$ based on GD is equivalent to optimizing parameters $w$ based on NGD.

\paragraph{Proof}  This equivalence is derived by applying the chain rule of derivatives and exploiting the symmetry of the Fisher information matrix. 

First, the standard update rule of NGD is simplified to:
\begin{equation}\label{eq3}
w\leftarrow w-\alpha \cdot G^{-1}\cdot \frac{\partial l}{\partial w}
\end{equation}
It is important to note that this equation assumes the existence of the inverse matrix $G^{-1}$ of the Fisher information matrix. However, in practice, the Fisher matrix is typically positive semi-definite and may not have a full-rank inverse. To address this issue, regularization or approximation techniques are commonly employed, which will be discussed in detail later.

Then, we multiply both sides of Eq. (\ref{eq3}) by $G^{\frac{1}{2}}$ and obtain:
\begin{equation}\label{eq4}
G^{\frac{1}{2}} \cdot w \leftarrow G^{\frac{1}{2}} \cdot w-\alpha \cdot G^{\frac{1}{2}} \cdot G^{-1} \cdot \frac{\partial l}{\partial w}
\end{equation}
\begin{equation}
\iff G^{\frac{1}{2}} \cdot w \leftarrow G^{\frac{1}{2}}\cdot w -\alpha \cdot G^{-\frac{1}{2}} \cdot \frac{\partial l}{\partial w}
\end{equation}

Next, apply the chain rule of derivatives to transform $\frac{\partial l}{\partial w}$ as follows:
\begin{equation}\label{eq5}
G^{\frac{1}{2}} \cdot w =G^{\frac{1}{2}} \cdot w -\alpha \cdot G^{-\frac{1}{2}} \cdot [\frac{\partial (G^{\frac{1}{2}} \cdot w)}{\partial w}]^T \cdot \frac{\partial l}{\partial (G^{\frac{1}{2}} \cdot w)}
\end{equation}

After taking the derivative $\frac{\partial (G^{\frac{1}{2}} \cdot w)}{\partial w}$ of the above equation, we get:
\begin{equation}\label{eq6}
G^{\frac{1}{2}} \cdot w \leftarrow G^{\frac{1}{2}}\cdot w -\alpha \cdot G^{-\frac{1}{2}} \cdot [G^{\frac{1}{2}}]^T \cdot \frac{\partial l}{\partial (G^{\frac{1}{2}} \cdot w)}
\end{equation}

Given that the Fisher information matrix $G$ is symmetric, we have $G^{\frac{1}{2}} = [G^{\frac{1}{2}}]^T$. Therefore, Equation (\ref{eq6}) can be simplified as:
\begin{equation}\label{eq7}
G^{\frac{1}{2}} \cdot w \leftarrow G^{\frac{1}{2}} \cdot w-\alpha \cdot G^{-\frac{1}{2}} \cdot G^{\frac{1}{2}}\frac{\partial l}{\partial (G^{\frac{1}{2}} \cdot w)}
\end{equation}
\begin{equation}
\iff G^{\frac{1}{2}} \cdot w \leftarrow G^{\frac{1}{2}} \cdot w-\alpha \cdot \frac{\partial l}{\partial (G^{\frac{1}{2}} \cdot w)}
\end{equation}

By substituting $w' = G^{\frac{1}{2}} \cdot w$ into the above equation, we obtain the following update rule, which is consistent with traditional gradient descent (GD):
\begin{equation}\label{eq8}
w'\leftarrow w'-\alpha\cdot \frac{\partial l}{\partial w'}
\end{equation}

Hence, optimizing $w$ using NGD is equivalent to optimizing $w'$ using GD.
\subsection{Proposed method}
\label{cal}

Capitalizing upon proposition 1 above, we proposed a novel optimization method called structured natural gradient descent (SNGD). More specifically, the boost in efficiency brought by our method is built upon two elements: 1) Decomposing the network into subsystems, and 2) Efficient computation of the square roots of the Fisher matrix $G$ addressing two computational challenges: 1) the cubic complexity involved in computing the Fisher matrix G, and 2) The matrix inversion. 

\subsubsection{System decomposition}
To address the challenge of computing the complex global Fisher information matrix during the training of deep neural networks with NGD, we employ a decomposition approach. Here, we partition the network into hierarchical subsystems, enabling the calculation of localized Fisher matrices, which can be calculated more efficiently.
\begin{figure}[t]
  \centering
  \includegraphics[width=\columnwidth]{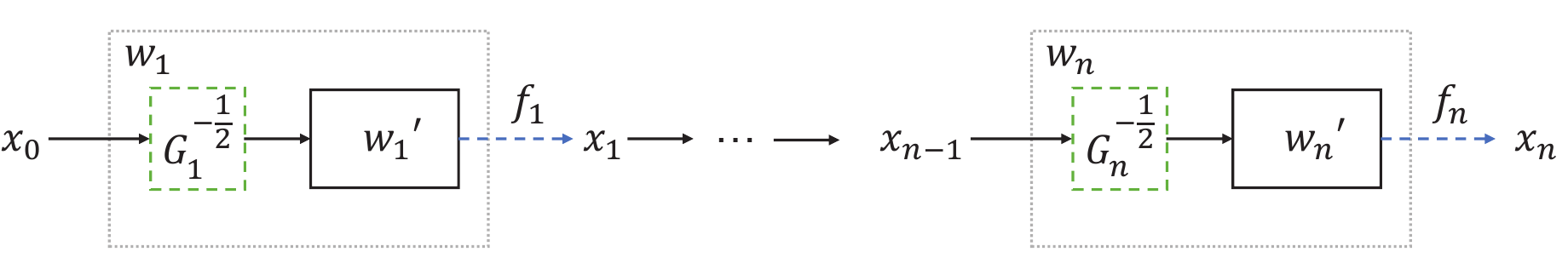}
  \caption{The restructured deep neural network.
  }\label{fig2}
\end{figure}
 Each subsystem represents a  parameterized network layer to which we append an additional sub-layer with a new normalization operator of $G^{-\frac{1}{2}}$, and new weight parameters $w'$, defined via the structural transformation over $w'$ (Eq.\ref{eq9}), and an activation function $f$. 
\begin{equation}\label{eq9}
w=G^{-\frac{1}{2}}\cdot w'
\end{equation}
This above transformation allows for optimizing $w'$ via traditional yet efficient GD optimization. 

Let us consider a $n$-layer deep neural network, as shown in Fig. \ref{fig2}. With the above decomposition, the network is composed of a sequence of subsystem layers. Specifically, the subsystem of the $k$-th layer ($k\in\{1,2,\dots,n\}$) consists of a normalization sub-layer with a value of the negative square root of local Fisher matrix $G_k^{-\frac{1}{2}}$, a sub-layer with new weight parameters $w_k'$, and an activation function $f_k$.

\subsubsection{Optimal computation of the square roots of Fisher matrix $G$}
The local Fisher information matrix for each layer is determined as outlined below \cite{sun2017relative}:
\begin{equation}\label{eq10}
    G\leftarrow {\rm E}(v_f){\rm E}(xx^T)
\end{equation}
where ${\rm E}(\cdot)$ denotes the expectation, $x$ represents the input of one layer, $f$ is the activation function of this layer, and $v_f$ is the derivative of the activation function, which means the effective learning area of the neuron.

The computational complexity of SNGD primarily lies in $G^{\frac{1}{2}}$ and $G^{-\frac{1}{2}}$. To mitigate this, we employ an iterative optimization technique.
In Eq. (\ref{eq10}), the main calculation cost lies in ${\rm E}(xx^T)$, which takes $O(n^3)$. As $xx^T$  in the neural network corresponds to the Gram matrix, we approximate it using the Nystrom method \cite{drineas2005nystrom}. This method adopts an iterative scheme to compute square roots $Z$ of $A$  (Eq. (\ref{eq11}),
\begin{equation}\label{eq11}
F(Z)\leftarrow Z^2-A=0
\end{equation}
Using the Denman-Beavers iteration method \cite{denman1976matrix} with initial values $Y_0=A$ and $Z_0=I$, the iterative operations are defined as follows:
\begin{equation}\label{eq12}
Y_{k+1}\leftarrow \frac{1}{2} (Y_k+Z_k^{-1}), Z_{k+1}\leftarrow \frac{1}{2}(Z_k+Y_k^{-1})
\end{equation}
Through Eq. (\ref{eq12}), the matrices $Y_k$ and $Z_k$ can quickly converge to $A^{\frac{1}{2}}$ and $A^{-\frac{1}{2}}$. To further improve the efficiency, we modify the iterative Eq. (\ref{eq11}) referring to the iterative method \cite{lin2017improved} to avoid inverse operations. The optimized iterative operations are as follows:
\begin{equation}\label{eq13}
Y_{k+1}\leftarrow \frac{1}{2} Y_k (3I-Z_k Y_k ),Z_{k+1}\leftarrow \frac{1}{2} (3I-Z_k Y_k ) Z_k
\end{equation}
By offering an approximate solution to computing the inverse of $G$, this method enables us to derive the negative square root of the matrix entirely via matrix multiplication which complexity can go down to $O(n^{2.373})$  as compared to $O(n^{3})$ for the matrix inversion \cite{gall2018}. 
This not only demands fewer memory and computational resources as a fundamental principle but also mitigates issues with numerical stability, particularly when the matrix is poorly conditioned.

\subsection{SNGD Training of a deep neural network}
After restructuring the network and adjusting its structure via the new normalization sub-layers, we use GD to optimize the new weight parameters, though retaining similar benefits as NGD. The computation of local Fisher layers in SNGD facilitates input regularization by leveraging $E(xx^T)$ and enables the identification of effective neurons by considering the variations in transformations through $E(v_f)$.
 These local Fisher layers keep the data distribution stable and provide a curvature signal of the loss of space. Therefore, this new method has a significant effect on both accelerating network convergence and regularization, making gradient updates more stable and faster.
 The optimization procedure of deep neural networks using our proposed SNGD is depicted in Algorithm 1.
 
\begin{algorithm}
\caption{SNGD on a  deep neural network}\label{alg:algorithm}
Initialize ${G_k}^{-\frac{1}{2}}$  as the identity matrix, \\
Iterate steps 1 to 4 in this way until convergence or termination of training.
\begin{algorithmic}[1]
\STATE Step 1: Calculate $G_k$, according to Eq. (\ref{eq10}), $G_k\leftarrow E(v_{f_k})E(x_k x_k^T )$;
\STATE Step 2: Use the efficient method described in Section \ref{cal} to calculate $G_k^{\frac{1}{2}}$ and $G_k^{-\frac{1}{2}}$;
\STATE Step 3: Calculate $w_k^{'}\leftarrow G_k^{\frac{1}{2}} \cdot w_k$;
\STATE Step 4: Use traditional GD to optimize $w_k^{'}$;
\end{algorithmic}
(Note: The $G_k^{-\frac{1}{2}}$, as shown in the green box in Fig. \ref{fig2}, does not participate in the gradient calculation of the backpropagation)
\end{algorithm}
\label{algorithm1}

\section{Experiments}
\subsection{Experimental settings}
\textbf{Networks.} We evaluate the proposed method on several widely used networks as below.
\begin{itemize}
\item MLP: The MLP \cite{krizhevsky2009learning} has two fully connected hidden layers, each with 80 hidden units and ReLU as activation functions. 
\item VGG: The VGGNet \cite{simonyan2014very} consists of convolution, pooling, and fully-connected layers. The kernel size is 3x3, and the max-pooling layer is of the size of 2x2 with a stride of 2. The activation functions after the convolution or fully-connected layer are all ReLU. VGG-16 is taken.
\item ResNet: ResNet-18 \cite{he2016deep} is taken.
\item 1, 2, 3-layer LSTM \cite{ma2015long} are both with 200 hidden units in each layer.
\end{itemize}

\textbf{Datasets and loss functions.} We explore image classification on different datasets, including MNIST \cite{lecun1998gradient} and CIFAR dataset \cite{krizhevsky2009learning}. The MNIST and CIFAR-10 datasets for image classification consist of 10 different classes, with 28x28 gray images and 32x32 color images respectively, divided into a training set of 50,000 images and a test set of 10,000 images. Besides, to evaluate whether SNGD can generalize well to training on larger image recognition datasets, we also train ResNet on ImageNet \cite{deng2009imagenet} which contains 5247 classes and 3.2 million images. As for LSTM, we perform language modeling on the Penn Treebank dataset \cite{marcinkiewicz1994building}, a corpus consisting of over 4.5 million words of American English. The loss functions are all cross-entropy loss functions based on $L_2$-regularized ($l_2=10^{-3}$).

\textbf{Compared optimizers.} Due to SNGD being essentially a first-order normalization method, we perform extensive comparisons with traditional first-order network optimizers, such as SGD \cite{sutskever2013importance}, Adam \cite{kingma2014adam}, and RAdam \cite{liu2019variance}. 
Additionally, we use the performance of KFAC \cite{martens2015optimizing} on MNIST as the baseline for the second-order optimizer to validate the effectiveness of our proposed SNGD. 
For these traditional first-order optimizers, we perform grid searches to choose the best hyperparameter and report the results using the best hyperparameter settings. 
For the learning rate, we search it among {100.0, 10.0, 1.0, 0.1, 0.01, 0.001, 0.0001}. 
For image classification experiments, we set the learning rate of 0.1 for SGD with momentum and SNGD, 0.001 for Adam and RAdam. 
For SGD, we set the momentum as 0.9, which is the default for many networks such as ResNet. 
For other parameters, we set them as their default values in the literature \cite{sutskever2013importance, kingma2014adam, liu2019variance, martens2015optimizing}.

\subsection{SNGD performance}
In this subsection, we verify the performance of SNGD for the image classification task on MNIST and show the results of SNGD against second-order baseline KFAC and first-order optimizer SGD on MNIST.

\begin{figure}
  \centering
  \includegraphics[width=\columnwidth]{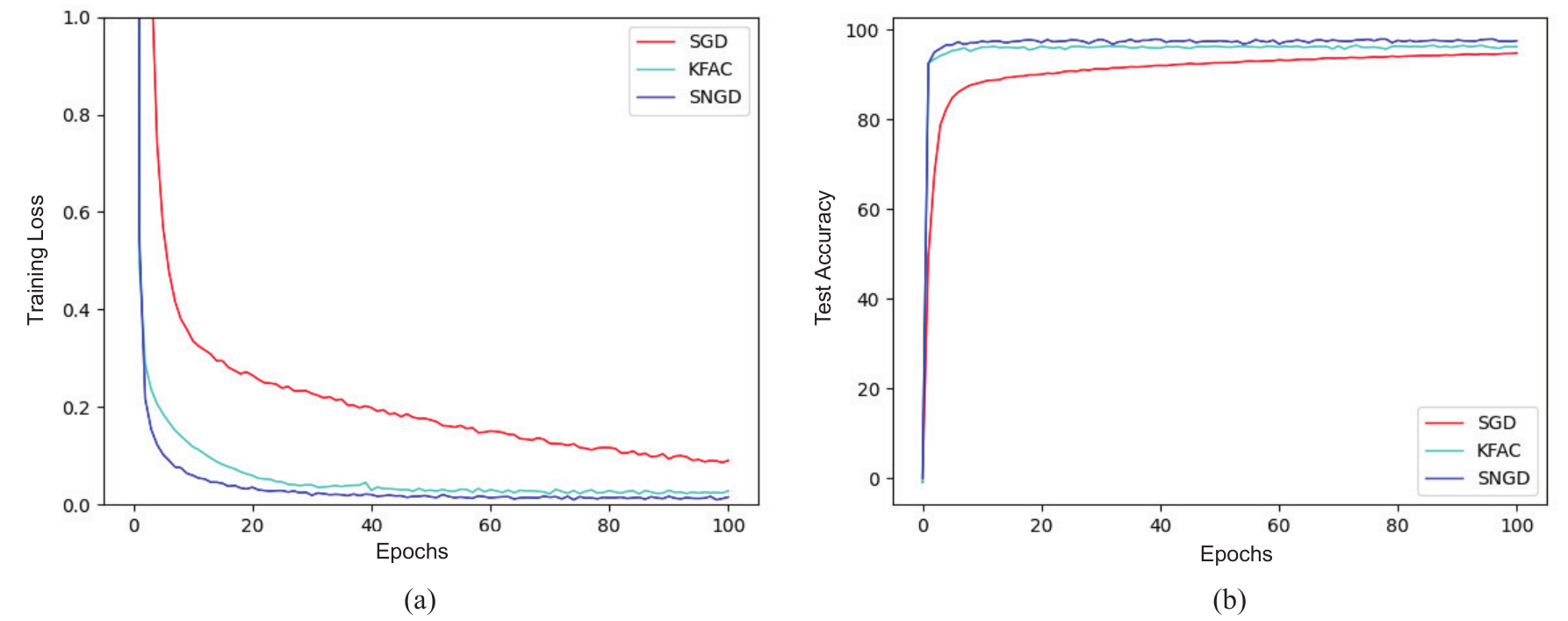}
  \caption{Performances of our method for image classification task on MNIST.}\label{fig3}
\end{figure}

Fig. \ref{fig3}(a) and (b) show the training loss and test accuracy for image classification task on MNIST, and we find that second-order algorithms, both SNGD and KFAC, perform far better than first-order optimizers SGD. 
From Fig. \ref{fig3}, we can see that our structured method converges faster than SGD while achieving higher test accuracy (97.6\%) than KFAC (96.3\%) and SGD (94.8\%). 
The performance of SNGD greatly exceeds the original first-order optimizer SGD thanks to the advantages of the second-order optimization algorithm. 
Furthermore, our method is a more precise implementation of NGD than KFAC and takes into account the computational efficiency of the Fisher matrix, which leads SNGD to surpass the second-order optimizer baseline. 
However, generalizing this second-order optimization method baseline to large-scale deep networks is difficult to realize. So, in the following subsection, we compare well-tuned first-order optimizers with SNGD in various tasks.

\subsection{Comparisons}
\subsubsection{Local Fisher layer versus batch norm layer}
\begin{figure}
  \centering
  \includegraphics[width=\columnwidth]{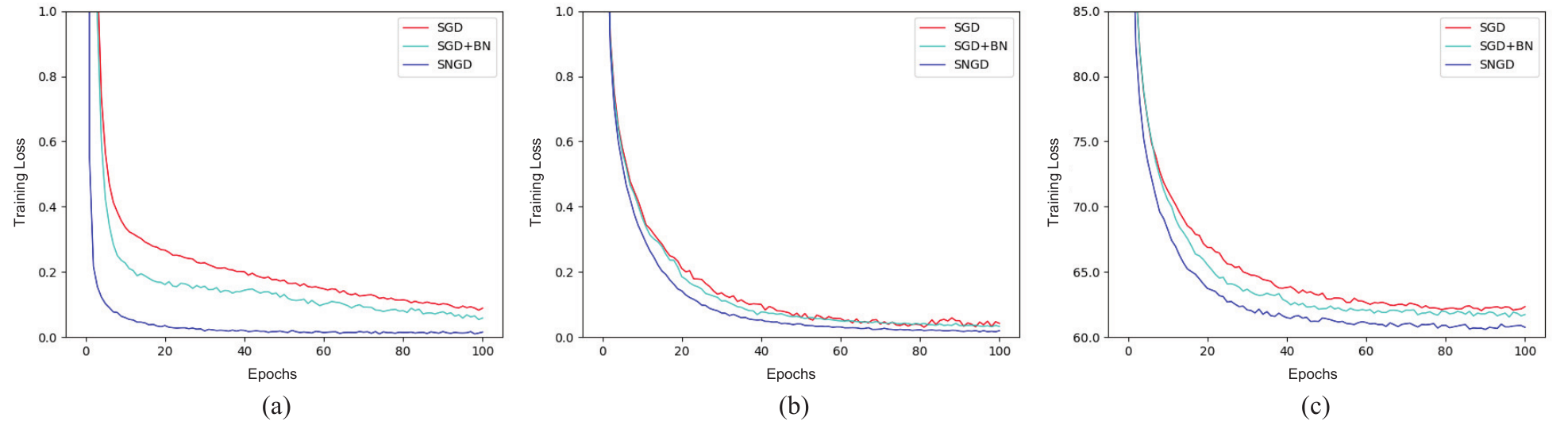}
  \caption{Training Loss Comparisons: Local Fisher Layer in SNGD vs. Batch Norm in SGD+BN vs. SGD without Batch Norm.}\label{fig4}
\end{figure}

To demonstrate the effectiveness of key components of SNGD, namely the Local Fisher layer, we compare the Local Fisher layer to the Batch Norm layer with the same model and hyperparameters (as detailed previously) on MNIST, CIFAR-10, and Penn Treebank datasets. In Fig. \ref{fig4}(a\textasciitilde c), we apply SNGD, SGD+BN, and SGD to train MLP on MNIST, VGG-16 on CIFAR-10, and 2-layer LSTM on Penn Treebank, respectively. For example, as shown in Fig. \ref{fig4}(a), the final loss of MLP trained with SNGD reaches 0.036 on the MNIST dataset, less than the final loss of 0.112 trained with SGD. Although the loss of SGD+BN is reduced to 0.094, it is still inferior to SNGD. Those results confirm the superiority of the proposed method and show that SNGD can accelerate the convergence of deep network models and achieve better performance than SGD while maintaining its computational simplicity. 

\subsubsection{Comparison with traditional first-order GDs}
We first demonstrate the effectiveness of SNGD in training different networks, including MLP, VGG-16, and ResNet, compared with traditional first-order optimizers, including SGD, Adam, and RAdam.

\begin{figure}
  \centering
  \includegraphics[width=\columnwidth]{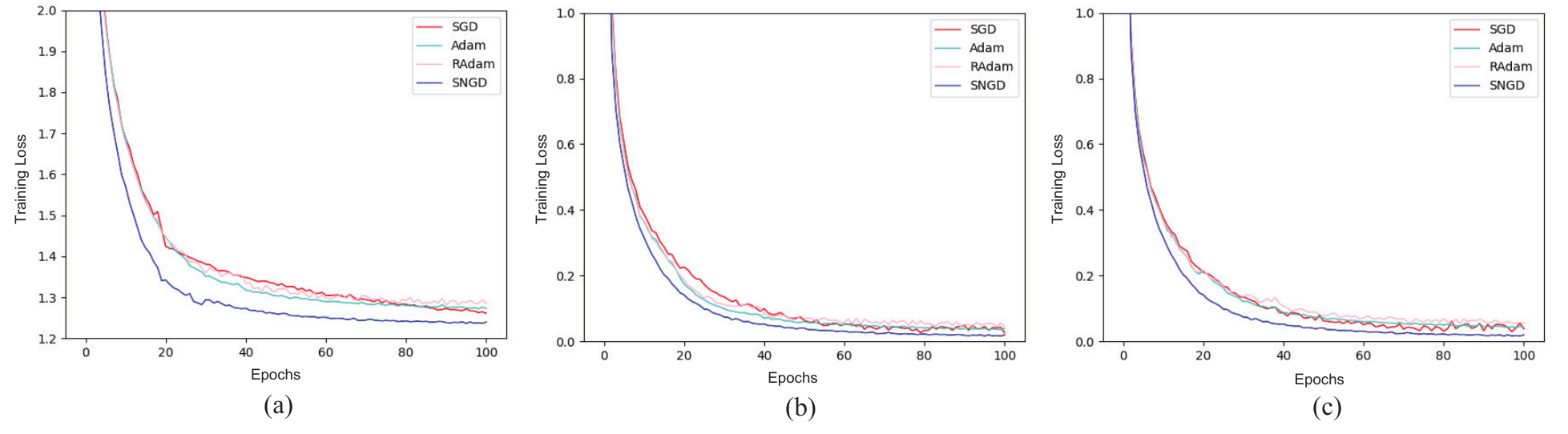}
  \caption{The training loss of different networks on CIFAR10.}\label{fig5}
\end{figure}

\textbf{MLP.} As shown in Fig. \ref{fig5}(a), SNGD significantly outperforms other optimizers trained for 100 epochs using an MLP with two hidden layers on the CIFAR-10 dataset. However, traditional first-order optimizers reduce the loss at early steps but saturate quickly.

\textbf{VGG-16.} Fig. \ref{fig5}(b) shows the training curves of traditional GD training VGG-16 on CIFAR-10 for 100 epochs. Overall, SNGD achieves the best performance with the lowest loss.

\textbf{ResNet.} Fig. \ref{fig4}(c) illustrates the training curve with the lowest loss when training ResNet-18 on CIFAR-10. This shows that SNGD also achieves the lowest loss on ResNet-18. These observations demonstrate the effectiveness of our SNGD approach, which generalizes better to training tasks than other stochastic first-order optimizers.

We further compare the training process of SNGD on the different datasets of 100 epochs and a batch size of 50 with SGD, Adam, and RAdam, respectively. 

\begin{figure}
  \centering
  \includegraphics[width=\columnwidth]{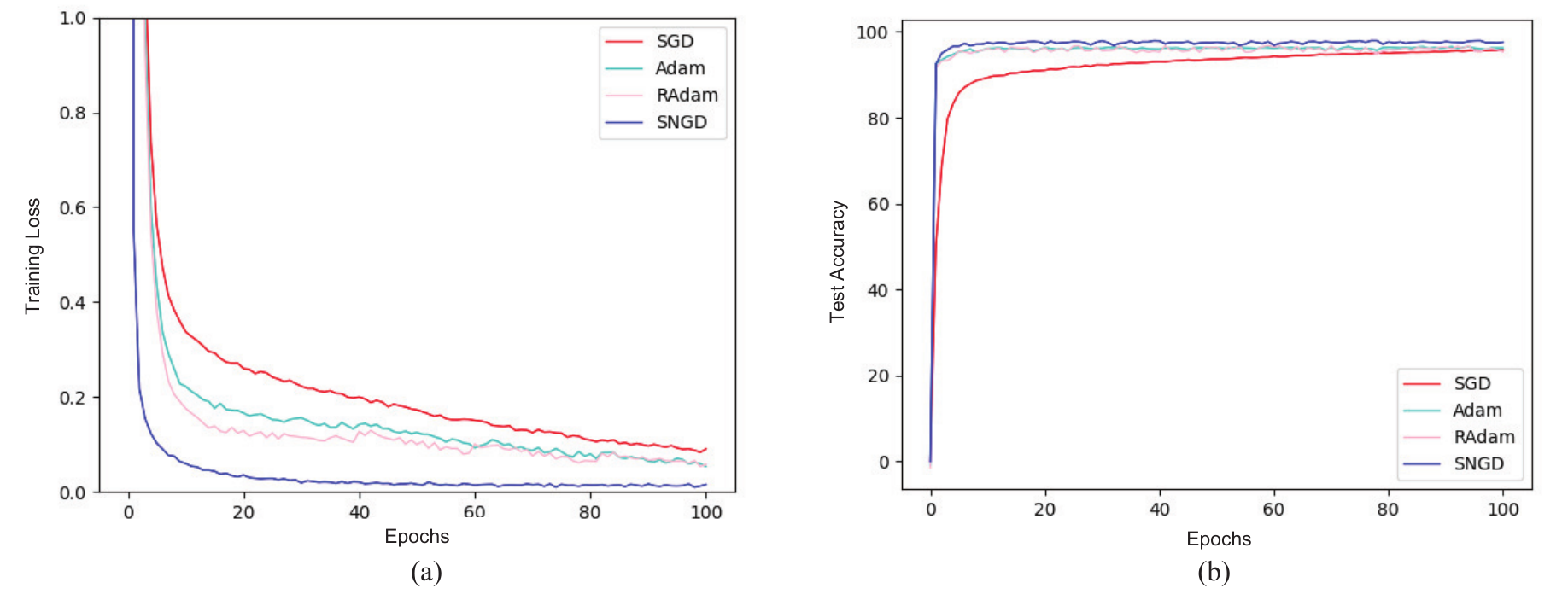}
  \caption{Training loss and test accuracy on MNIST by training MLP for 100 epochs.}\label{fig6}
\end{figure}

\textbf{MNIST.} We train MLP on the MNIST dataset for 100 epochs, and the comparison of training and testing curves using different optimizers are shown in Fig. \ref{fig6}. The results show that SNGD significantly outperforms traditional stochastic first-order optimizers and adaptive methods, including SGD, Adam, and RAdam. 

\textbf{CIFAR.} As shown in Fig. \ref{fig5}, we compare the effectiveness of SNGD with other stochastic first-order methods in training MLP, VGG-16, and ResNet-18 on CIFAR-10. SNGD achieves a comparable convergence speed as adaptive methods such as Adam while achieving better accuracy than SGD and other methods without incurring excessive time cost.

\begin{figure}
  \centering
  \includegraphics[width=\columnwidth]{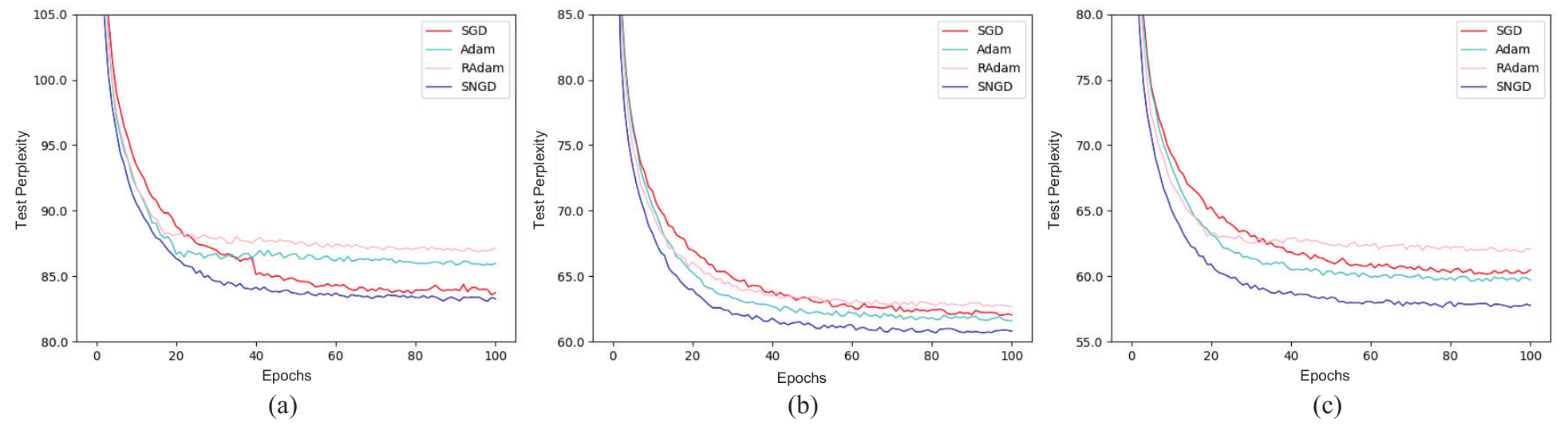}
  \caption{Test perplexity on Penn Treebank for 1,2,3-layer LSTM. Lower is better.}\label{fig7}
\end{figure}

\textbf{Penn Treebank.} We conduct experiments with a 1,2,3-layer LSTM with several hidden units of 200 and report the perplexity on the test set trained with different optimizers in Fig. \ref{fig7} and Table \ref{table1}. We can observe that on the simpler 1,2-layer LSTM model, the difference between different optimizers is not very obvious. But different optimizers have different optimization behaviors on more complex 3-layer LSTM models. Although Adam and RAdam converge slightly faster in the early stages, they quickly fall into local optimal solutions. At the same time, SNGD can quickly update the optimization direction, thus achieving the best final test complexity in language modeling tasks.

\begin{table}
  \caption{Test Perplexity Across 1, 2, 3-Layer LSTM Models on the Penn Treebank Dataset: Emphasizing Lower Scores for Superior Performance with Best results Highlighted in Bold.}
  \label{table1}
  \centering
  \begin{tabular}{@{}lllll@{}}
\toprule
    Model & SNGD & SGD & Adam & RAdam \\
\midrule
    1-layer LSTM & \textbf{83.92} & 84.21 & 86.43 & 88.74\\
    2-layer LSTM & \textbf{61.58} & 63.37 & 61.58 & 63.87\\
    3-layer LSTM & \textbf{58.27} & 61.22 & 60.44 & 63.32 \\
\bottomrule
\end{tabular}
\end{table}

\subsection{Generalization of the trained networks on test sets}
\begin{figure}
  \centering
  \includegraphics[width=\columnwidth]{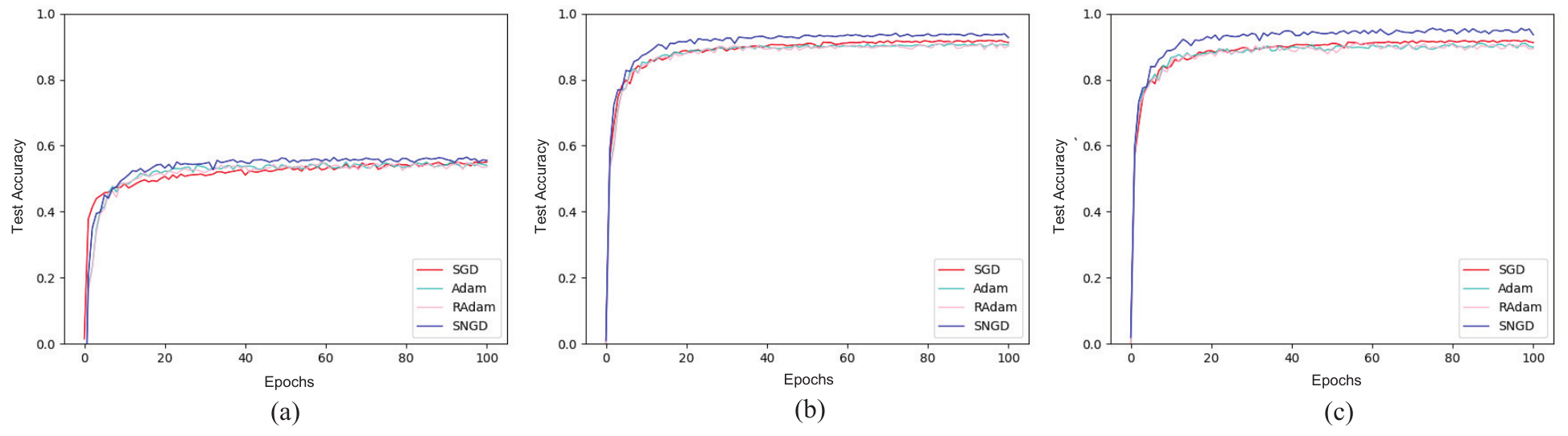}
  \caption{Comparison of SNGD with other optimizers for training MLP, VGG-16, and ResNet-18 on CIFAR10.}\label{fig8}
\end{figure}

We further evaluate the generalization ability of the trained MLP, VGG-16, and ResNet-18 on CIFAR10 with different optimizers. The Top-1 accuracy when the network converges is reported in Table \ref{table2}. Fig. \ref{fig8} shows the highest accuracy verified. According to Table \ref{table2}, MLP, VGG-16, and ResNet-18 trained by SNGD all achieve the best Top-1 accuracy on the test set. These results show that SNGD is beneficial to improve the generalization ability of the trained network.

\begin{table}
  \caption{Top-1 accuracy of MLP, VGG-16, ResNet-18 on test set of CIFAR-10. Best results are in bold.}
  \label{table2}
  \centering
\begin{tabular}{@{}llll@{}}
\toprule
    Optimizer & MLP & VGG-16 & ResNet-18 \\
\midrule
    SGD & 54.37\% & 92.64\% & 93.02\% \\
    Adam & 53.86\% & 92.12\% & 92.93\% \\
    RAdam & 53.92\% & 92.16\% & 92.91\% \\
    SNGD & \textbf{55.57}\% & \textbf{94.03\%} & \textbf{94.44\%} \\
\bottomrule
\end{tabular}
\end{table}

According to Fig. \ref{fig5}(a\textasciitilde c) on the training dataset, SGD and SNGD are the top two best optimizers. On the test dataset, our SNGD-trained model performs best, as shown in Table \ref{table2}. We consistently observe that the model trained by Adam achieves the best Top-1 accuracy on the training set but underperforms on the test set, according to Table \ref{table2}. On the test set, ResNet-18 trained with Adam achieves 93.02\% Top-1 accuracy, lower than 94.44\% trained with SNGD. Our SNGD-trained models generalize consistently well on the test set, thanks to the regularization and nonlinear transformation of our local Fisher layers subtly providing a curvature signal of the loss space. 

To further test the stability of our method, we then train a ResNet-18 for 100 epochs on ImageNet and report the Top-1 accuracy on the test set in Table \ref{table3}. For other optimizers, we report the best results given in the literature \cite{chen2018closing} and \cite{liu2019variance}. SNGD surpasses alternative optimization methods, demonstrating superior accuracy compared to traditional gradient descent and adaptive methods, particularly on larger datasets. These findings validate the exceptional generalization performance of SNGD.

\begin{table}
  \caption{Top-1 accuracy of ResNet18 on ImageNet. * is reported in \cite{chen2018closing}, ** is reported in \cite{liu2019variance}}
  \label{table3}
  \centering
\begin{tabular}{@{}llll@{}}
\toprule
    SNGD & SGD & Adam & RAdam \\
\midrule
    73.41\% & 70.23\%* & 63.79\%* & 67.62\%** \\
\bottomrule
\end{tabular}
\end{table}

\subsection{Computational time}
To evaluate the computational efficiency of our proposed SNGD method, we compare it with traditional first-order optimizers using a 6-core Intel i9-8950HK CPU and an NVIDIA GeForce GTX 1080 GPU. Table \ref{table4} shows the training time for MLP, VGG-16, and ResNet-18 with a batch size of 50 on CIFAR-10 per epoch using SNGD and traditional first-order optimizers. Our results indicate that SNGD achieves comparable training time to traditional first-order optimizers while exhibiting superior performance, as demonstrated in Table \ref{table4}. We attribute this to the fact that most first-order optimizers, such as Adam, update parameters by computing the parameter update vector and assigning it to the network layer by layer. At the same time, our method employs the Gram matrix and Newton's method to accelerate the computation of the negative square root of the local Fisher information matrix, thereby retaining the natural gradient's accelerated convergence properties.

\begin{table}[h]
  \caption{Comparison of runtime. The values represent the time for training per epoch on CIFAR-10 with batch size 50.}
  \label{table4}
  \centering
\begin{tabular}{@{}llll@{}}
\toprule
    Optimizer & MLP & VGG-16 & ResNet-18 \\
\midrule
    SGD & 4.96s & 38.38s & 52.89s \\
    Adam & 5.02s & 38.42s & 52.94s \\
    RAdam & 4.97s & 38.38s & 52.87s \\
    SNGD & 5.35s & 39.52s & 53.81s \\
\bottomrule
\end{tabular}
\end{table}

\subsection{Resource Utilization and Efficiency}
In this subsection, we further provide an empirical comparison of Gradient Descent (GD), Natural Gradient Descent (NGD), and our proposed Structured Natural Gradient Descent (SNGD) across multiple models and datasets. We analyze these optimization methods with respect to convergence time (Conv. Time), memory consumption (Mem), and GPU/CPU utilization, both average (Avg. Util) and peak (Peak. Util). Table \ref{table5} presents a synthesized overview of the comparative metrics.

\begin{table}[h]
  \caption{Comparative Analysis of Resource Utilization and Efficiency in GD, NGD, and SNGD.}
  \label{table5}
  \centering
  \small 
  \begin{tabular}{@{}lcccc@{}}
  \toprule
    Model/Dataset & Metric & GD & NGD & SNGD \\
  \midrule
    \multirow{4}{*}{MLP/MNIST} 
    & Conv. Time (s) & 5 & 6 & 5.3 \\
    & Mem. (GB) & 1 & 1.5 & 1.2 \\
    & Avg. Util. (\%) & 39 & 32 & 35 \\
    & Peak Util. (\%) & 50 & 45 & 48 \\
  \addlinespace
    \multirow{4}{*}{VGGNET34/CIFAR} 
    & Conv. Time (s) & 27 & 53 & 32 \\
    & Mem. (GB) & 3 & 5 & 3.9 \\
    & Avg. Util. (\%) & 62 & 48 & 57 \\
    & Peak Util. (\%) & 90 & 82 & 88 \\
  \addlinespace
    \multirow{4}{*}{RESNET34/IMAGENET} 
    & Conv. Time (s) & 23 & 49 & 37 \\
    & Mem. (GB) & 2.9 & 4.8 & 4.2 \\
    & Avg. Util. (\%) & 59 & 51 & 55 \\
    & Peak Util. (\%) & 91 & 81 & 86 \\
  \addlinespace
    \multirow{4}{*}{LSTM/MNIST} 
    & Conv. Time (s) & 18 & 35 & 26 \\
    & Mem. (GB) & 1.3 & 2.7 & 1.8 \\
    & Avg. Util. (\%) & 46 & 38 & 42 \\
    & Peak Util. (\%) & 75 & 69 & 74 \\
  \bottomrule
  \end{tabular}
\end{table}

As shown in Table \ref{table5}, the proposed SNGD demonstrates notable improvements in convergence times compared to NGD across all tested models. This increase in efficiency is especially significant in complex models such as VGGNET34/CIFAR and RESNET34/IMAGENET, where the high dimensionality of parameters can substantially slow down the optimization process. SNGD shows an advantageous balance, converging faster than NGD while avoiding the extensive memory utilization associated with this method. In terms of memory consumption, SNGD operates with a moderate overhead compared to GD but remains less demanding than NGD. This outcome indicates SNGD's capability to maintain a relatively low memory footprint while still benefiting from the structured approach to natural gradients. GPU/CPU utilization metrics provide insights into the computational behavior of SNGD. While average utilization for SNGD does not significantly deviate from GD and NGD, peak utilization suggests that SNGD can occasionally reach similar levels of resource demand as GD. This characteristic may imply that SNGD can intensify computational efforts when necessary, possibly contributing to its faster convergence times without sustaining high utilization throughout the optimization process.\\

The structured approach taken by SNGD, therefore, offers a compelling trade-off between computational efficiency and the convergence properties offered by NGD. By effectively managing the trade-offs between memory and computation, SNGD emerges as a robust alternative that leverages the benefits of both traditional GD and NGD. Future work should focus on validating these benefits in a broader range of applications and establishing the practical limits of the SNGD approach.

\section{Conclusions}

This paper has presented a novel optimization framework for training deep neural networks with Structured Natural Gradient Descent (SNGD). By reconstructing the network layer in a deep neural network, SNGD realizes natural gradient descent, accelerating network convergence and improving optimization effectiveness. Experimental results demonstrate that SNGD not only outperforms traditional GD optimization methods regarding convergence speed and accuracy on various training tasks, including MLP on MNIST, MLP, CNN, ResNet on CIFAR-10, ResNet on ImageNet, and LSTM on Penn Treebank, but also showcases its universal applicability and efficiency in resource utilization. Particularly, SNGD's capability to achieve faster convergence times across all tested models, as highlighted in our comparative metrics, underlines its superior efficiency. By maintaining lower memory consumption and dynamically managing GPU/CPU utilization, SNGD illustrates a significant advancement in optimizing deep learning computations. We believe that the SNGD method has the potential to expand the scope of natural gradient descent to encompass a wider array of deep networks, such as transformers, further enhancing its utility and effectiveness in the field of machine learning.



{\bibliographystyle{IEEEtran}

\bibliography{mybibliography}}

\end{document}